\documentclass[sigconf,nonacm]{acmart}

\usepackage{booktabs} 
\usepackage{marvosym}

\usepackage[ruled]{algorithm2e} 

\SetAlFnt{\small}
\SetAlCapFnt{\small}
\SetAlCapNameFnt{\small}
\SetAlCapHSkip{0pt}
\IncMargin{-\parindent}

\setcopyright{none}

\acmDOI{}

\setcitestyle{square}

\begin{document}
\title{AvatarBooth: High-Quality and Customizable \\ 3D Human Avatar Generation}

\author{Yifei Zeng, Yuanxun Lu, Xinya Ji, Yao Yao, Hao Zhu$\textsuperscript{\Letter}$, Xun Cao}

\affiliation{%
  \institution{Nanjing University}
  \city{Nanjing}
  \country{China}
}

\renewcommand{\shortauthors}{ }
\renewcommand{\shorttitle}{ }

\newenvironment{figurehere} 
    {\def\@captype{figure}} 
    {} 
\makeatother

\begin{abstract}

We introduce AvatarBooth, a novel method for generating high-quality 3D avatars using text prompts or specific images. Unlike previous approaches that can only synthesize avatars based on simple text descriptions, our method enables the creation of personalized avatars from casually captured face or body images, while still supporting text-based model generation and editing. 
Our key contribution is the precise avatar generation control by using dual fine-tuned diffusion models separately for the human face and body. This enables us to capture intricate details of facial appearance, clothing, and accessories, resulting in highly realistic avatar generations.
Furthermore, we introduce pose-consistent constraint to the optimization process to enhance the multi-view consistency of synthesized head images from the diffusion model and thus eliminate interference from uncontrolled human poses.
In addition, we present a multi-resolution rendering strategy that facilitates coarse-to-fine supervision of 3D avatar generation, thereby enhancing the performance of the proposed system. 
The resulting avatar model can be further edited using additional text descriptions and driven by motion sequences. Experiments show that AvatarBooth outperforms previous text-to-3D methods in terms of rendering and geometric quality from either text prompts or specific images. 

\end{abstract}


\begin{teaserfigure}
  \includegraphics[width=1.0\textwidth]{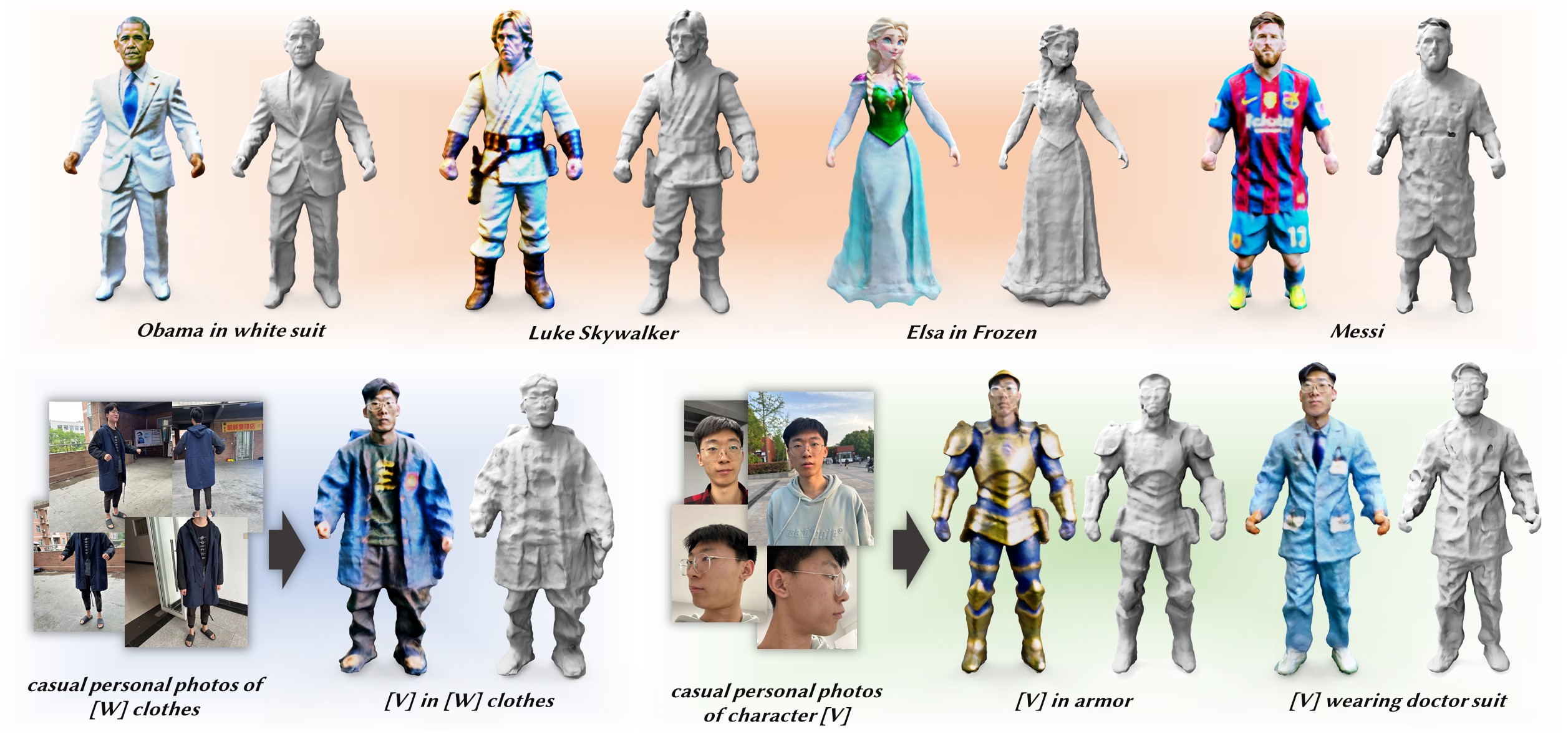}
  \vspace{-0.25in}
  \caption{We present AvatarBooth, a framework for generating 3D avatars from text prompts or certain images. Our method can generate 3D human avatars in \textit{prompt generative mode} (\textcolor[RGB]{132,60,12}{red}), \textit{appearance customized mode} (\textcolor[RGB]{31,78,121}{blue}), or \textit{hybrid mode} (\textcolor[RGB]{56,87,35}{green}). 
}
  \label{fig:teaser}
\vspace{0.15in}
\end{teaserfigure}

\maketitle

\section{Introduction}
\label{sec:intro}

Creating 3D human avatars from texts or images is a longstanding challenging task in both computer vision and computer graphics, which is key to a broad range of downstream applications including the digital human, film industry, and virtual reality. 
Previous approaches have relied on expensive and complex acquisition equipment to reconstruct high-fidelity avatar models~\cite{alexander2010digital, guo2017real, xiao2022detailed}. However, these methods require multi-view images or depth maps that are unaffordable for consumer-level applications. Alternatively, other methods leverage a neural network to predict plausible avatar models from a single image input~\cite{saito2019pifu, zheng2021pamir, xiu2022icon}. Nonetheless, these approaches are limited by the availability of suitable images and are non-editable once a reference image is provided.

Recently, 3D content generation based on large-scale pre-trained vision-language models has shown promising performance~\cite{poole2022dreamfusion, lin2022magic3d, raj2023dreambooth3d}. Specifically, these methods leverage the general 2D image priors learned from large-scale pre-trained models to guide the optimization of an implicit 3D representation. 
In early attempts, the contrastive language-image pre-training (CLIP)~\cite{radford2021clip} is leveraged to synthesize the appearance of the avatar given a text prompt~\cite{hong2022avatarclip, youwang2022clipactor}. Then, the Score Distillation Sampling (SDS)~\cite{poole2022dreamfusion} is further proposed to boost the performance by distilling the 2D knowledge from a pre-trained diffusion model~\cite{rombach2022high, ho2020denoising, saharia2022photorealistic} to 3D content generation via differentiable rendering. Although significant progress has been made, current methods are still unable to synthesize high-quality shapes and appearances of the human object, which contains complex poses and detailed 3D structures like cloth wrinkles and facial shapes.

On the other hand, generating a customized avatar of arbitrary identity that corresponds to input images remains a challenging problem. Though DreamBooth3D~\cite{raj2023dreambooth3d} provided a solution for generating personalized 3D assets, it struggles to reproduce the high-fidelity human face with the exact identity shown in images. A novel fine-tuning strategy is required to support both detailed appearance synthesis and multimodal-driven customization of the avatar.

In this paper, we propose a novel method, named \textit{AvatarBooth}, for generating high-quality and customizable avatars from text prompts or image sets. Our method aims to generate identity-customized 3D avatars that accurately reflect the visual and textual features of a specific individual. 
To this end, a neural implicit surface~\cite{wang2021neus} is learned to represent the shape and appearance of the human avatar, which is supervised with dual pre-trained or fine-tuned latent diffusion models for the face and body respectively. 
Meanwhile, the pose-consistent constraint is introduced to enhance the fine-tuning of the diffusion models in the task of appearance-customized generation, which provides more accurate multi-view supervision with a consistent appearance in a canonical pose space.
Furthermore, a multi-resolution SDS scheme is introduced to predict the fine structure and appearance of the avatar in a coarse-to-fine manner.

By leveraging a few pictures of a person, the model can synthesize 3D avatars that not only possess the individual's unique appearance but also match the abstract features specified in the input text prompt. These abstract features include attributes such as `wearing glasses or hats in a certain style', which are user-friendly in editing and modifying the avatar's overall visual identity. Our approach is designed to leverage priors in both large language-vision models as well as concrete input images, resulting in avatars that are faithful to the input appearance while also being editable through text prompts.

The contribution of this paper can be summarized as:

\begin{itemize}
    \item We propose a 3D human avatar generation framework that supports both text prompts and arbitrary images as input. Dual latent diffusion models are introduced to supervise the face and body generation separately, yielding detailed facial appearance, clothes, and wearings.
    
    \item Pose-consistent constraint is introduced to customize the large pre-trained diffusion models given photos of a specific person. We use ControlNet~\cite{zhang2023adding} to enhance the multi-view consistency of the synthesized images, so as to eliminate the interference of uncontrolled human poses and lead to high-quality appearance and geometry.
    
    \item We present a multi-resolution score distillation sampling strategy that supervises the generation of the 3D avatar in a coarse-to-fine manner. Experiments show that this strategy not only enhances the rendering quality but also improves the robustness of generation.
    
\end{itemize}

\section{Related Works}
\label{sec:related}

\begin{figure*}[t]
  \centering
  \includegraphics[width=1.0\linewidth]{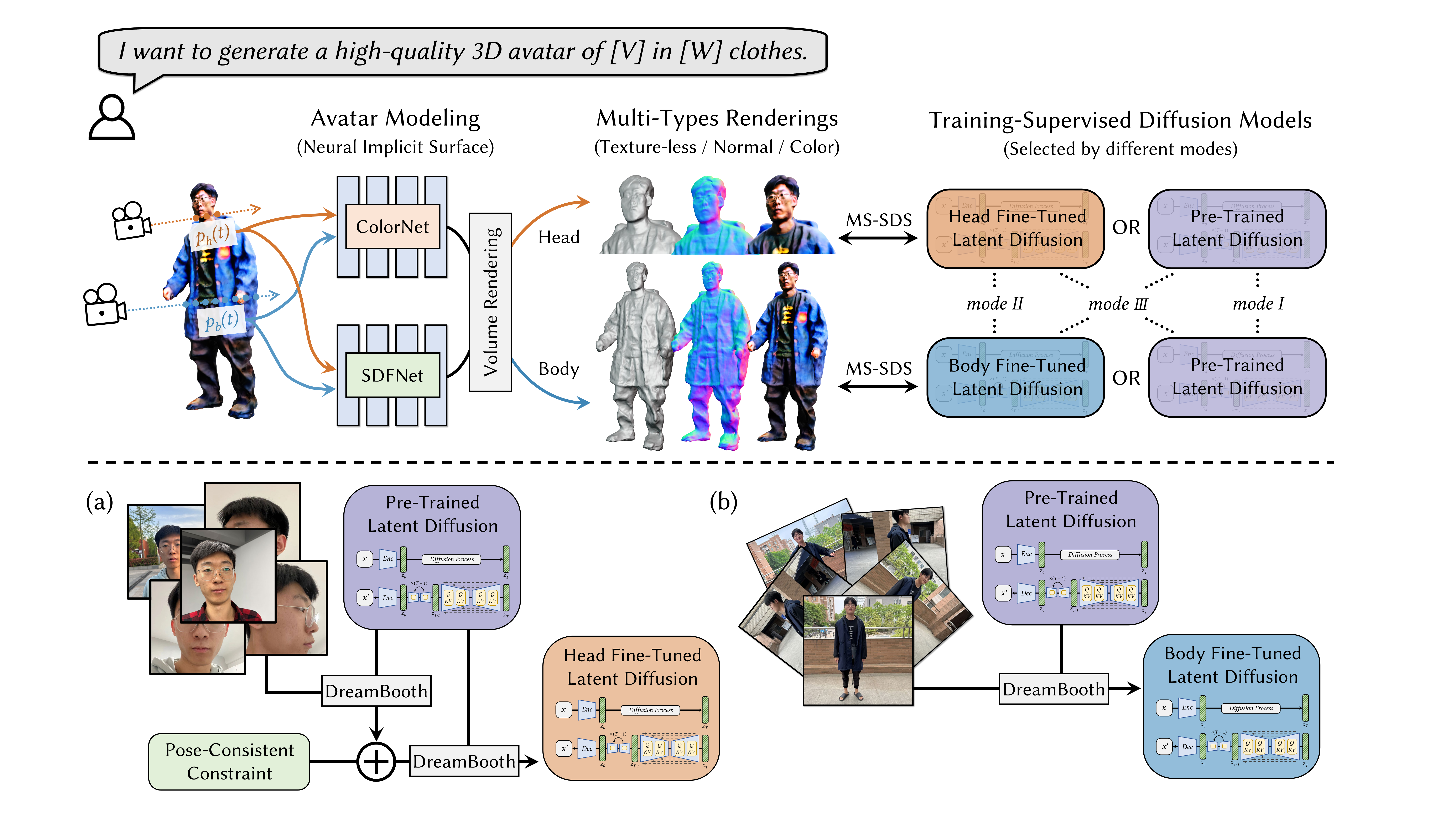}
  \vspace{-0.3in}
  \caption{\textbf{Overall pipeline.} Our method represents a human avatar with NeuS, which is initialized with an SMPL shape. Through the volume renderer, the avatar model is transformed into textureless, color, and normal renderings, which are used for SDS training with the supervision of diffusion models. By adopting a selection or combination of pre-trained and fine-tuned diffusion models, our approach can be performed in three modes: (I) prompt generative mode; (II) appearance customized mode; and (III) hybrid mode. The fine-tuning strategies for the face and full body are shown in (a) and (b) respectively.}
  \vspace{-0.1in}
\label{fig:pipeline}
\end{figure*}

\textbf{Text-guided 2D\&3D Generation.}
In recent years, diffusion models~\cite{ho2020denoising, song2020denoising, dhariwal2021diffusion} have rapidly developed due to their remarkable performance in synthesizing high-quality images. A core structure of diffusion models consists of forward diffusion steps that add noise according to a scheduler and backward generative steps that denoise the noise. In addition to unconditional generation from Gaussian noise only, the diffusion model can generate high-quality images from text prompts or images as input. Among the various diffusion models, the latent diffusion model~\cite{rombach2022high} has emerged as a promising text-to-image model that strikes a good balance between image quality and memory usage.
When it comes to 3D content generation, existing methods~\cite{poole2022dreamfusion, lin2022magic3d, chen2023fantasia3d} leveraged pre-trained text-to-image diffusion models to supervise coordinate-based networks with score distillation loss (SDS)~\cite{poole2022dreamfusion}. 
Other methods~\cite{wu2023high, wang2023rodin, zhang2023dreamface} leverage the 3D parametric face model~\cite{yang2020facescape, zhu2021facescape}, large pre-trained language-vision models~\cite{radford2021learning, rombach2022high}, or large-scale synthetic data~\cite{wood2021fake} to achieve the text-guided generation of high-fidelity 3D human faces.
Moreover, some approaches~\cite{metzer2022latent, seo2023ditto} adopt a neural radiance field to represent the latent space of Stable Diffusion, which enables the synthesis of novel views from text descriptions. Except for using SDS for guidance, other methods~\cite{wang2022score, hong2023debiasing} also use Score Jacobian Chaining to generate 3D assets with text, which takes into consideration the gradient of diffusion models. While these methods have successfully generated view-consistent 3D models, generating 3D avatars remains a challenging task due to the complexities involved in articulated 3D shape and appearance diversity.

\textbf{Finetuning of Diffusion Models.}
In recent years, with the growing interest in the text-to-image domain, pioneer researchers have explored personalizing text-to-image models using photos of specific subjects. One representative work is DreamBooth~\cite{ruiz2023dreambooth}, which leverages a rare token to represent a particular subject or style while preventing overfit with a prior preservation loss. Starting from another strategy, textual inversion~\cite{gal2022image} creates a new embedding for the input concept and optimizes this embedding vector with a few photos to achieve subject-driven image generation. LoRA~\cite{hu2021lora} proposes to fine-tune large language models, which freezes the pre-trained model weights and meanwhile injects learnable rank decomposition matrices into the layers of the Transformer network~\cite{vaswani2017attention}. LoRA significantly improves the ease of diffusion model fine-tuning by reducing the number of trainable parameters for downstream tasks.
DreamBooth3D~\cite{raj2023dreambooth3d} proposes a 3-stage iterative approach to produce 3D models using the fine-tuned personalized diffusion model with DreamBooth. However, DreamBooth3D fails to recover an identity-consistent and detailed human avatar, meanwhile, the generated personalized content is not editable based on text prompts. Therefore, we explore addressing these limitations and enhancing the flexibility of avatar generation based on both image sets and text prompts.

\textbf{Avatar Generation Models.}
Traditional methods for generating 3D avatars often rely on training on 3D datasets, which can be difficult to collect and scale up. To overcome this challenge, recent methods have utilized cheaper 2D data to train a neural field, such as EG3D~\cite{chan2022efficient}, GNARF~\cite{bergman2022generative}, EVA3D~\cite{hong2022eva3d}, HumanGen~\cite{jiang2022humangen}, ENARF-GAN. Additionally, some explicit methods~\cite{varol2018bodynet, zhu2016video, zhu2018view, zhu2019detailed, zheng2019deephuman, zhu2021detailed, alldieck2019tex2shape, xiu2023econ, han2023high} as well as implicit methods~\cite{saito2019pifu, saito2020pifuhd, huang2020arch, tan2020self, he2021arch++, zheng2021pamir, peng2021neural, xiu2022icon} have been developed to generate human avatars conditioned on a single input image. These methods have limitations in generating avatars with unseen styles, wearings, and appearances that are not present during training, let alone personalizing avatars with a certain identity via word descriptions.
AvatarCLIP~\cite{hong2022avatarclip} was the first to generate and animate 3D avatars in a zero-shot text-driven manner, while CLIP-Actor learns a displacement map from CLIP for mesh deformation and vertex coloring. More recently, following works like AvatarCraft~\cite{jiang2023avatarcraft} and DreamAvatar~\cite{cao2023dreamavatar} have utilized diffusion models to produce high-quality 3D avatars. However, AvatarCraft is limited in producing large deformations to the original template model, and their geometry has some artifacts in places like the chest and back. DreamAvatar generates high-quality clothing geometry and texture, but it is unable to extract a fine-grained mesh due to the inherent feature of Latent-NeRF, resulting in non-articulated results after the training process is finished. Our method manages to eliminate such issues and can synthesize a customized avatar from free-view images, which can be edited through text prompts.

\section{Method}
\label{sec:method}

\subsection{Preliminaries}

\textbf{Neural Implicit Surfaces (NeuS)}~\cite{wang2021neus}. 
NeuS is a neural implicit representation that represents a 3D surface as the zero-level set of a signed distance function (SDF)~\cite{park2019deepsdf}.
 
Given a coordinate $(x,y,z)$ and viewing location/direction $(o,d)$, two MLPs are used for predicting the SDF value and the RGB value respectively. Then, pixel colors can be calculated using the volume rendering equation:

\vspace{-0.1in}

\begin{equation}
C(o,d) = \sum\limits_{p \in R}^n{(w(t)c(p(t),d))}
\end{equation}

\vspace{-0.05in}

\noindent where $p(t)$ is a sampled point, and $R$ contains $n$ sampled points along the ray $o+t\cdot d$, and $w(t)$ is formulated as:

\begin{equation}
    w(t)=\frac{\phi_s(f(p(t)))}{\int_0^{\infty} \phi_s(f(p(u))) d u}
\end{equation}

\noindent where $\phi_s$ is the logistic density distribution, $f$ is an SDF network. Compared to NeRF~\cite{mildenhall2020nerf} and its variances, the formulation of NeuS eliminates bias in the first order of approximation, leading to more detailed surface reconstruction.

\textbf{Score Distillation Sampling (SDS)}~\cite{poole2022dreamfusion}. SDS is a strategy to generate subjects from a diffusion model by optimizing a loss function, which can be used to optimize an avatar represented by a 3D field. Specifically, the gradient of this score function indicates a higher density region for rendered images. The detailed formula is introduced in Section~\ref{sec:framework}.

In order to achieve personalized generation under the SDS strategy, DreamBooth3D proposes to supervise the learning of the subject's appearance with a diffusion model fine-tuned by DreamBooth~\cite{ruiz2023dreambooth}. In our method, we follow the SDS strategy cooperated with fine-tuned diffusion models, then made improvements to further improve the customizable capability and the accuracy of the synthesized appearance.

\subsection{Pipeline Overview}
\label{sec:framework}

Our method takes a set of images or text prompts as input and synthesizes a 3D detailed avatar represented by NeuS. As shown in Fig.~\ref{fig:pipeline}, the entire generation pipeline consists of three modules. In the avatar modeling module, a bare rendering of SMPL model~\cite{loper2015smpl, pavlakos2019smplx} is trained into a neural implicit field~\cite{wang2021neus} that consists of an SDF network $f(x; \theta)$ and a color network $c(x; \theta)$, following prior works~\cite{hong2022avatarclip, jiang2023avatarcraft}. 

In the rendering module, three types of renderings are obtained from pre-defined virtual cameras located around the avatar space. Empirically, we rendered normal maps $\mathcal{I}_n$ in addition to color  and texture-less renderings $\{\mathcal{I}_c, \mathcal{I}_g\}$, and experiments demonstrate that the introduction of normal maps enhances the geometric details such as facial contours and cloth wrinkles. Then, we leverage the SDS Loss to guide the NeuS to converge, which can be formulated as:

\vspace{-0.05in}

\begin{equation}
    \nabla \psi \mathcal{L}_{\mathrm{SDS}}(\phi, \mathcal{I})=\mathbb{E}\left[w(t)\left(\hat{\epsilon}_\phi\left(z_t^{\mathcal{I}} ; y, t\right)-\epsilon\right) \frac{\partial \mathcal{I}}{\partial \psi} \frac{\partial z^{\mathcal{I}}}{\partial \mathcal{I}}\right]
\end{equation}

\noindent where, $\phi$ represents the parameters of the diffusion model, $\mathcal{I}$ is the image used for supervision including $\{\mathcal{I}_g,\mathcal{I}_n,\mathcal{I}_c\}$, and $z^{I}$ is the corresponding latent code of the image $\mathcal{I}$. The function $\epsilon()$ represents the noise predicted by the diffusion model, while $y$ and $t$ denote the input prompt and timestep, respectively.
To optimize the face and human body simultaneously, we adopted two sets of rendering parameters centered on the face and the whole human body respectively, which will be detailed in Section~\ref{sec:two_part}. 

In the SDS training modules, pre-trained and fine-tuned latent diffusion models are selected or combined to supervise the training of NeuS via the renderings. A multi-resolution training of SDS is implemented to model the avatar in a coarse-to-fine manner, which will be detailed in Sec.~\ref{sec:multi_res}. In the fine-tuning of the latent diffusion models, we propose to introduce the pose-consistent constraint, which will be detailed in Sec.~\ref{sec:pose_consis}. According to how the pre-trained diffusion models are used in SDS training, our framework may work in the following three modes:

    \textbf{(I) Prompt generative mode.} Similar to AvatarCLIP~\cite{hong2022avatarclip} and AvatarCraft~\cite{jiang2023avatarcraft}, we use only text prompts as input to generate avatars that conform to the description without fine-tuning the pre-trained diffusion models. Since text prompts can only describe general or well-known appearances, this mode only works for synthesizing avatars with roughly matched appearances or celebrities. 
    
    \textbf{(II) Appearance customized mode.} We propose to customize the diffusion models as well as the learned avatars to match the appearance given a set of images. These images can be full-body or facial images taken freely from any viewpoint. Details of the appearance and clothing are passed on to generate the avatar model, even if the input picture contains only an incomplete or slightly contradictory appearance.
    
    \textbf{(III) Hybrid mode.} The above two modes can be performed simultaneously in a single model generation, that is, the hybrid mode. This mode can achieve the more complex conditional generation of an avatar, such as modifying the subject's clothes, hairstyle, age, beard, etc., through text prompts on the premise of synthesizing appearance according to input images.

\subsection{Dual Model Fine-tuning}
\label{sec:two_part}

We propose to leverage two diffusion models to supervise the training of the whole body and head, and the two models are also fine-tuned respectively. Though the previous works~\cite{hong2022avatarclip, jiang2023avatarcraft} augment the rendering samples around the face to improve the facial details, they do not exploit the potential of the fine-tuned vision-language models, so their attempts cannot enhance the performance of personalized avatar generation.

We initially uses only one diffusion model to supervise the training of the full body. We observe that a single SDS loss with the fine-tuning strategy of DreamBooth3D fails to strike a balance between the modeling of the facial appearance and the body clothes. Specifically, in early training steps, the appearance of clothes on the body is learned but the facial appearance is still unclear. If more training steps are made, the facial appearance will turn clear, but the global features like clothes style may be overfitted to the input images, which means it is hard to edit the body via text prompts in the hybrid mode. Besides, we also observed that Img2Img stage of DreamBooth3D can't produce accurate character identity faithful to the input images. We believe that this is due to the large difference in the scale between facial appearance and body appearance, which leads to the inconsistent convergence rate in the SDS training.

To address this issue, we propose the dual model fine-tuning strategy. When running in appearance customized mode or hybrid mode, the input images are divided into full body shots and headshots, which are used for fine-tuning two pre-trained models respectively. In the SDS training phase, we randomly sample cameras around the face and the whole body, then employ different diffusion models to guide the generation of the face and body, using head-oriented rendering and full-body rendering respectively. In the fine-tuning of the head shots model, we also introduce the pose-consistent constraint, which will be detailed in Sec.~\ref{sec:pose_consis}.

\subsection{Pose-Consistent Constraint}
\label{sec:pose_consis}

\begin{figure}[t]
  \centering
  \includegraphics[width=1.0\linewidth]{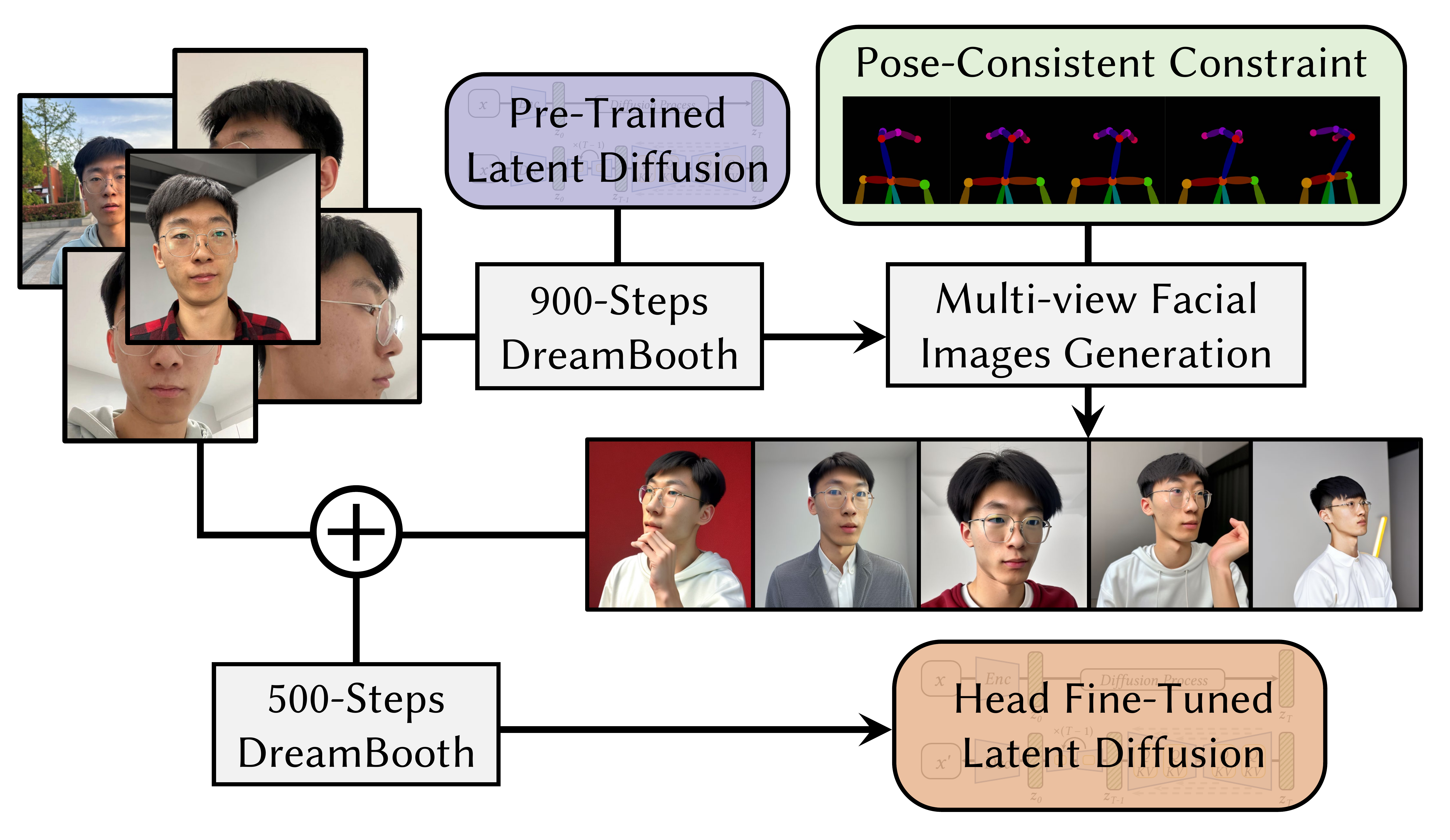}
  \vspace{-0.3in}
  \caption{\textbf{Pose-Consistent Constraint.} We first train a initial DreamBooth model $\hat{D}_{init}$ using input images $\mathcal{I}_{real}$ for plenty of steps. Guided by pose constraint rendered from Openpose, we can produce multi-view facial images $\mathcal{I}_{mv}$ that share the same identity with the person in the input images. Then, we combine the multiview images $\mathcal{I}_{mv}$ along with the input personal images $\mathcal{I}_{real}$ to fine-tune the final DreamBooth $\hat{D}_{final}$ for more steps.}
  \vspace{-0.1in}
\label{fig:face_fine-tune}
\end{figure}

To enhance the facial details of avatars generated from fine-tuned diffusion models, we propose a pose-consistent fine-tuning method by introducing ControlNet~\cite{zhang2023adding}. Previous approaches\cite{raj2023dreambooth3d} proved that directly utilizing DreamBooth with SDS-based methods will result in unsatisfactory outcomes, as the DreamBooth model tends to overfit the camera views used during fine-tuning.

In this work, we propose a two-stage strategy that utilizes ControlNet to incorporate more facial prior to the training process. Specifically, we first train an initial DreamBooth model $\hat{\mathcal{D}}_{init}$ using the input images $\mathcal{I}_{real}$. Then, we employ a keypoint ControlNet to produce multiview facial images $\mathcal{I}_{mv}$ guided by a skeleton constraint generated by OpenPose~\cite{cao2021openpose, cao2017realtime}, which are rendered in surround views. These synthetic images  $\mathcal{I}_{mv}$ are then combined with real images $\mathcal{I}_{real}$ to further fine-tune a new diffusion model $\hat{\mathcal{D}}_{final}$ by the DreamBooth method, thereby augmenting the facial details of the 3D model. 
Unlike previous methods\cite{raj2023dreambooth3d} that attempt to solve this issue during the training process of NeRF, our approach leverages ControlNet to address this problem before training a Neural Surface Field. As a result, we can use the same DreamBooth model $\hat{\mathcal{D}}_{final}$ to generate different avatars with the same identity without re-training from scratch. Experiments show that the use of ControlNet to guide the generation of multi-view facial images in combination with the DreamBooth model leads to more accurate and realistic avatars. 

\subsection{Multi-Resolution SDS}
\label{sec:multi_res}

As directly rendering high-resolution images from neural implicit filed is very computationally expensive, a common solution is to render a low-resolution image, then up-sample it to a higher resolution for SDS training~\cite{lin2022magic3d, chen2023fantasia3d}. The up-sampled images are then encoded to the latent space and used to supervise the training of a neural implicit field. However, we observed that increasing the upsampled resolution directly can lead to training collapse or inconsistent appearance.

To address these issues, we propose a multi-resolution optimization strategy, which gradually improves the up-sampling resolution for more stable SDS training. Starting from $H \times W$ images $\{\mathcal{I}_g,\mathcal{I}_n,\mathcal{I}_c\}$ rendered from NeuS, we initialize the network by training an up-sampled resolution of $512 \times 512$ for a few steps, then gradually improve the supervision resolution to $640 \times 640$ and $768 \times 768$. The lower resolution in the early training steps provides a coarse but robust starting point for the training process, while the higher resolution in the latter steps helps learn detailed geometry and high-quality appearance. Through experiments, we demonstrate that this simple strategy efficiently improves the stability in the early training stage and augments the appearance quality, ultimately yielding a more accurate and visually plausible avatar. 

\section{Experiments}
\label{sec:exp}

\begin{figure*}[ht]
  \includegraphics[width=0.9\textwidth]{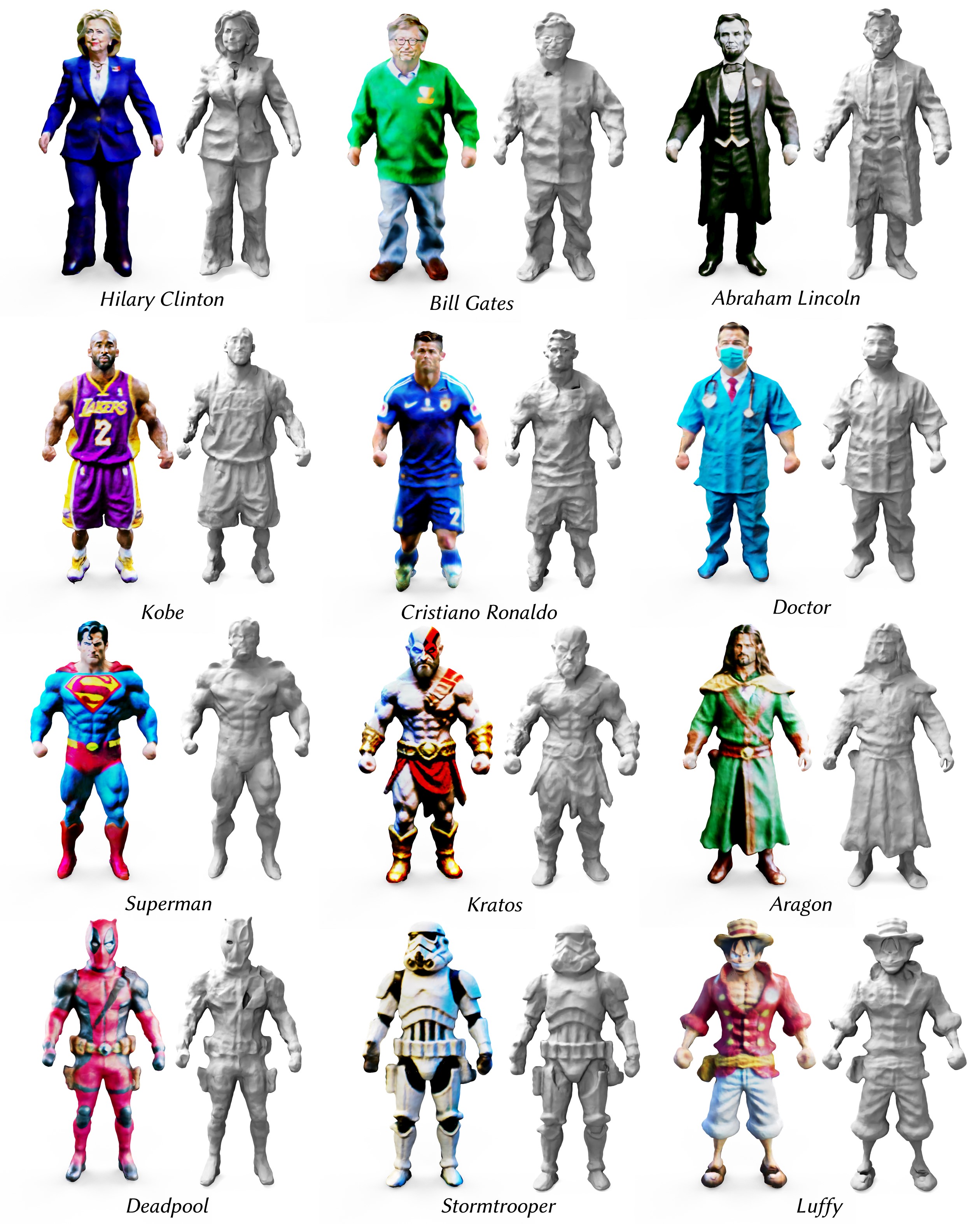}
  \caption{\textbf{Results in prompt-generative mode.} Our method recovers fine geometric shapes and textures, and the resulting human avatars closely match the text prompts.}
  \label{fig:exhibit_mode1}
\end{figure*}

\begin{figure*}[ht]
  \centering
  \includegraphics[width=1.0\linewidth]{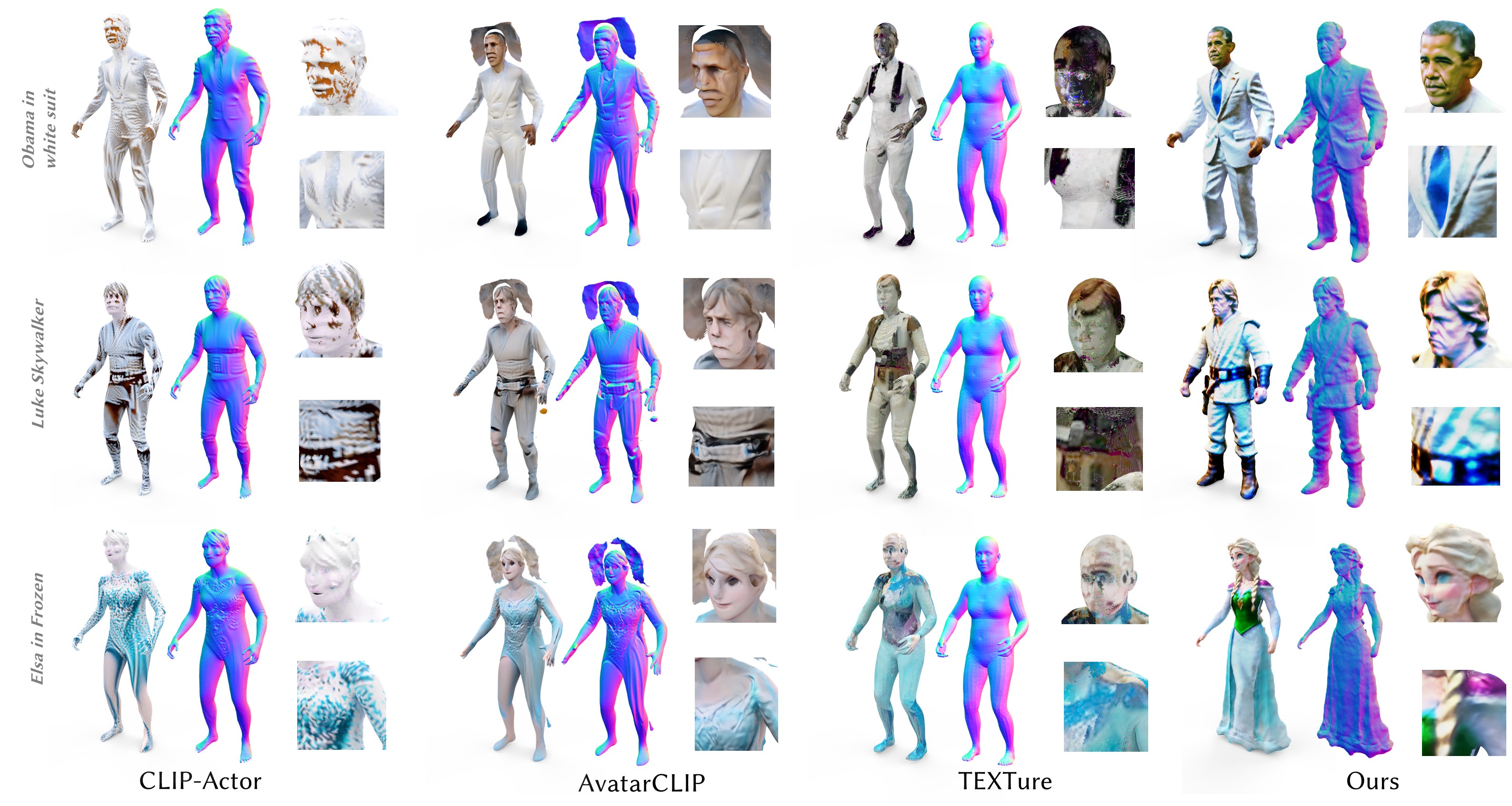}
  \vspace{-0.3in}
  \caption{\textbf{Qualitative comparisons-I.} We visualize the avatars generated by ours and previous works by rendering the model into color images and normal maps. The models generated by our method contain better geometric details, such as clothing lines and facial features, and have better rendering quality.}
  \vspace{-0.1in}
\label{fig:comparison}
\end{figure*}

In this section, we verify the effectiveness of the proposed method through experiments and compare and discuss it with previous methods. 

\subsection{Implementation details}

To model the neural implicit surface, we use a 6-layer MLP for the SDF network and a 4-layer MLP for the color network. 
To generate an avatar, we train the network for 8000 steps, including 2000 steps under an interpolation resolution of $512 \times 512$, 2000 steps under an interpolation resolution of $640 \times 640$, and 4000 steps under an interpolation resolution of $768 \times 768$. In the diffusion model fine-tuning phase, we train the first DreamBooth model for $900$ iterations to produce multi-view images, then use the generated images combined with the personal images to train a second DreamBooth model for $500$ steps.
We randomly sample the virtual camera locations for rendering, which contains $25\%$ centering on the face and the other $75\%$ centering on the overall body. The normal maps, shadow images, and color images are randomly rendered at a ratio of 1:1:8. These hyper-parameters are the same in all three modes.  
Adam optimizer~\cite{kingma2015adam} is used to train our model and the learning rate is set to $0.005$. On an NVIDIA RTX3090 GPU, it takes about 90 minutes to synthesize the avatar model and another 30 minutes to complete the fine-tuning of the diffusion model if appearance-customized generation is required. 

\subsection{Qualitative Evaluation}

\begin{figure}[t]
  \includegraphics[width=1.0\linewidth]{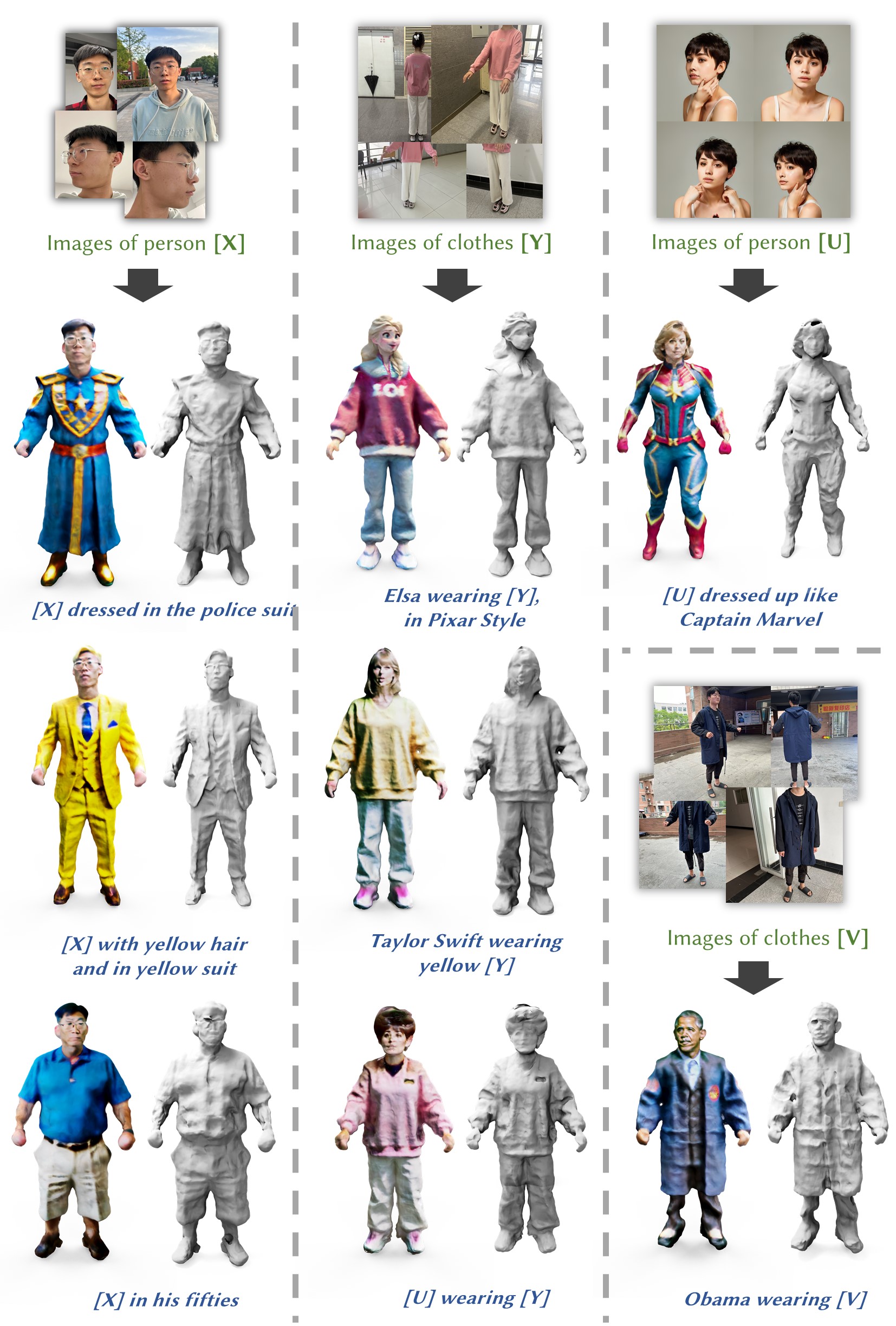}
  \caption{
  \textbf{Results in appearance customized / hybrid mode.} We can see that the human appearance from input image sets is transferred to the generated avatar, even if the images are captured in free conditions. The customized avatar can be further modified according to the additional text prompts.}
  \label{fig:exhibit_mode23}
\end{figure}

Our results of prompt generative mode are shown in Fig.~\ref{fig:exhibit_mode1}. We can see that our method synthesizes plausible human avatars with detailed geometry and fine appearance, which closely match the input text prompts. Our results of appearance customized mode and hybrid mode are shown in Fig.~\ref{fig:exhibit_mode23}. In these experiments, we can see that the human appearance from input image sets is transferred to the generated avatar, even if the images are captured in free conditions. The customized avatar can be further modified according to the additional text prompts. For example, by using simple prompts like `'a [V] man with yellow hair', the yellow hair will appear on the result avatar's head accordingly. Moreover, we demonstrated that even more abstract prompts like '[V] wearing like a wizard' and '[V] in his fifties' were effective, meanwhile, the customized appearance from the input image set is maintained if working in the hybrid mode. By employing text prompts, we were able to produce a diverse range of avatars with different appearances and styles, providing users with an efficient and personalized way of creating their desired avatars. Our result models can be easily rigged for animation. We recommend watching the video for more results and surrounding-view rendering.

\begin{figure}[t]
  \centering
  \includegraphics[width=1.0\linewidth]{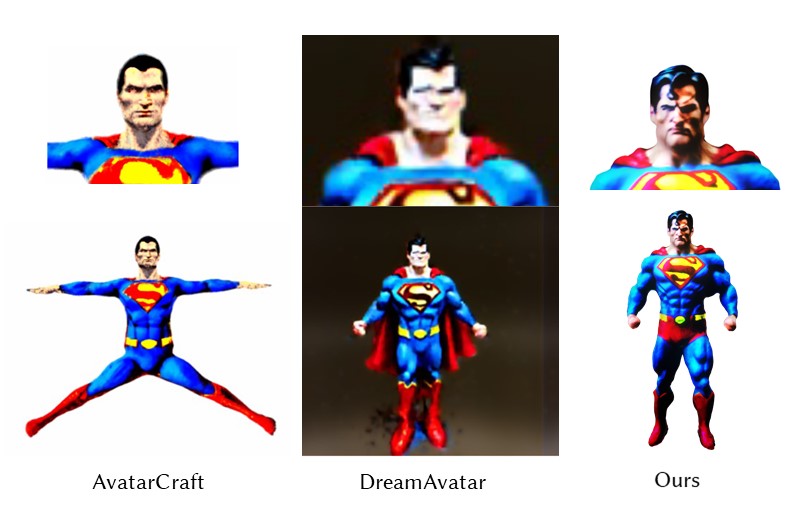}
  \caption{\textbf{Qualitative comparison-III.} We compare our method with AvatarCraft and DreamAvatar using the rendered results provided by the authors.}
  \label{fig:comparision2}
\end{figure}

We also compare our results qualitatively with prior works. Considering that previous approaches do not support customized avatar generation based on image sets, we only compare our methods in prompt generative mode with previous text-to-avatar methods, as shown in Fig.~\ref{fig:comparison}. As there are no official implementations for AvatarCraft~\cite{jiang2023avatarcraft} and DreamAvatar~\cite{cao2023dreamavatar}, we compare the performance with the same setting in their experiments, as shown in Fig.~\ref{fig:comparision2}. Our model achieves significant improvements over existing approaches in terms of both geometry and appearance. Specifically, our method can generate high-quality geometry and textures while preserving the character's identity. It is worth noting that, our method enables the generation of avatars with loose garments and accessories, which cannot be achieved by CLIP-Actor~\cite{youwang2022clipactor}, AvatarCLIP~\cite{hong2022avatarclip}, TEXTure~\cite{richardson2015texture}, AvatarCraft~\cite{jiang2023avatarcraft}. We believe that this is due to the fact that previous methods rely on the constraints from SMPL~\cite{loper2015smpl, pavlakos2019smplx} to maintain the avatar generation, while our method efficiently utilizes the vision-language prior and can generate a more generalized avatar model without heavy constraints from SMPL shape. DreamAvatar~\cite{cao2023dreamavatar} is also free of the SMPL constraints, however, the appearance and geometry of its resulting model are less detailed than that of our results.

\begin{figure}[t]
  \centering
  \includegraphics[width=1.0\linewidth]{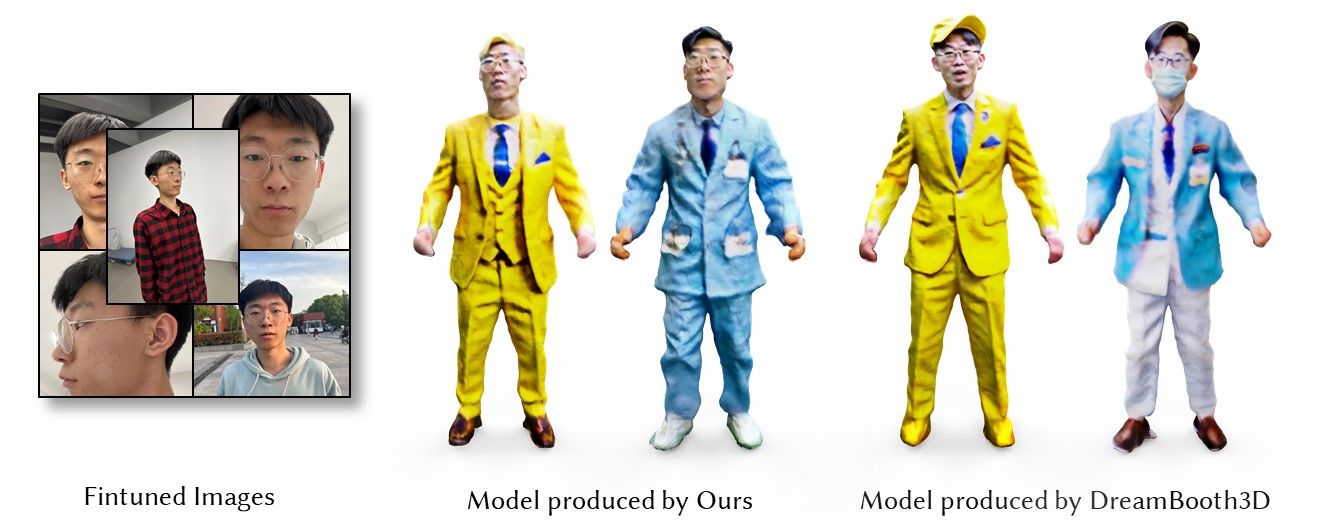}
  \caption{\textbf{Qualitative comparison-II.} We compare our method with DreamBooth3D in appearance-customized mode and hybrid mode.}
  \label{fig:personal}
\end{figure}

We trained NeuS following the strategy of DreamBooth3D  with the images above, and we find that DreamBooth3D can't supervise Neural Surface Field to produce 3D assets consistent with the input images. This can be attributed to the inaccuracy of facial details generated from the Img2Img stage in DreamBooth3D. Simply changing the prompt. characters will share different identities in DreamBooth3D results. 
Compare with DreamBooth3D, our method produces results more faithful to the fine-tuned images. Besides, our results will keep the same identity under all kinds of textual descriptions. 

\subsection{Quantitative Evaluation}

\begin{figure}[t]
  \centering
  \includegraphics[width=1.0\linewidth]{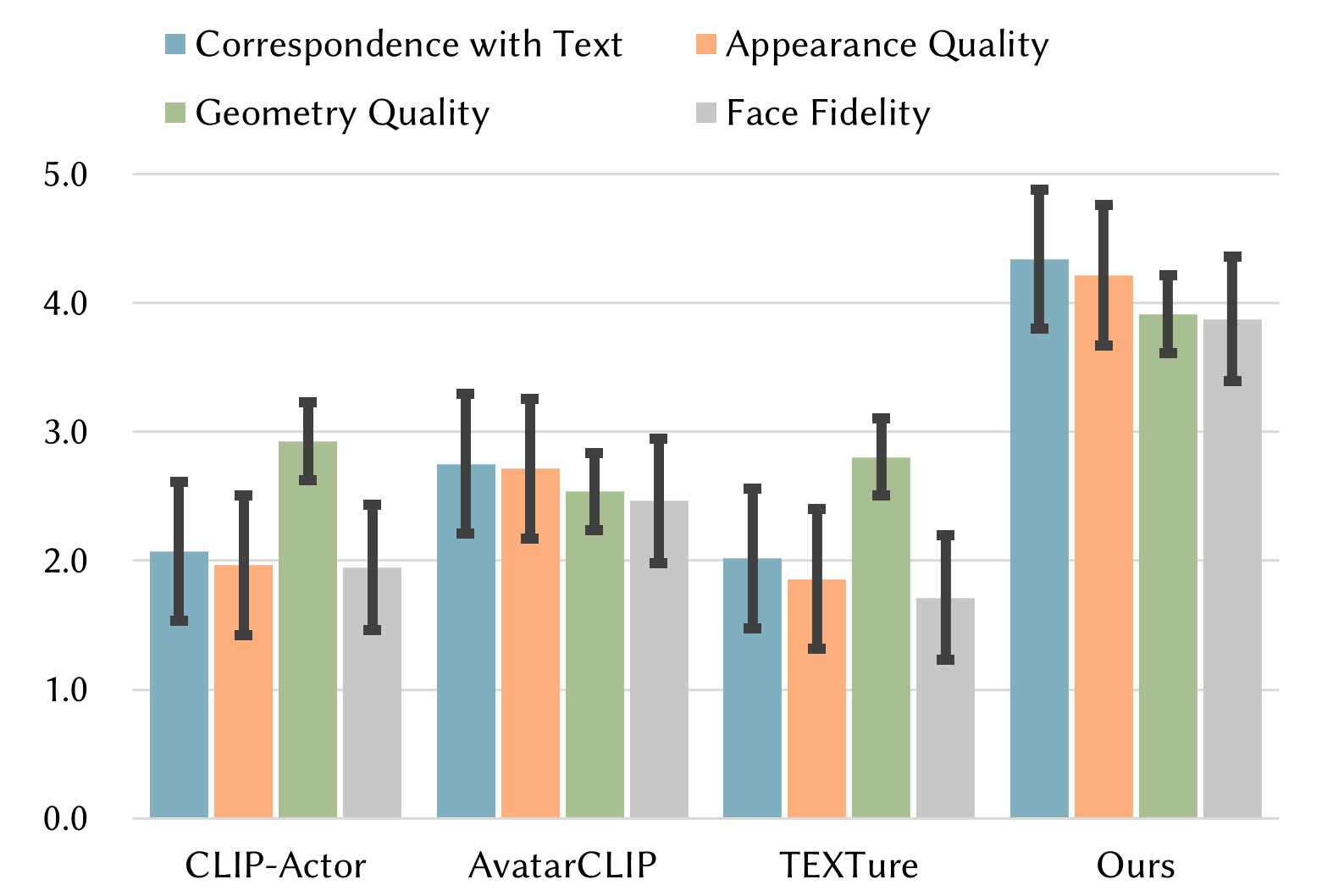}
  \vspace{-0.3in}
  \caption{\textbf{User Study.} We investigated the user's evaluation of our method and previous works w.r.t. correspondence with the text prompts, appearance quality, geometry quality, and face fidelity. Our method achieves optimal evaluation in all four metrics.}

  \label{fig:userstudy}
\end{figure}

\textbf{User Study.} To quantify the quality of the synthesized avatars, we conduct a user study to compare our results with generated ones from other state-of-the-art methods, \textit{i.e.} CLIP-Actor, AvatarCLIP and TEXTure. We generate 10 avatars from randomly selected text prompts for each method and recruit 30 volunteers to evaluate the results w.r.t four different aspects: correspondence with the text prompts, appearance quality, geometry quality, and face fidelity. The volunteers are required to score from 1 (worst) to 5 (best) for each term. The results are shown in Fig.~\ref{fig:userstudy}. Our method achieves the highest score over all four aspects, which proves that we are capable of generating avatars with more detailed appearance and geometry. 

\begin{figure}[t]
  \centering
  \includegraphics[width=1.0\linewidth]{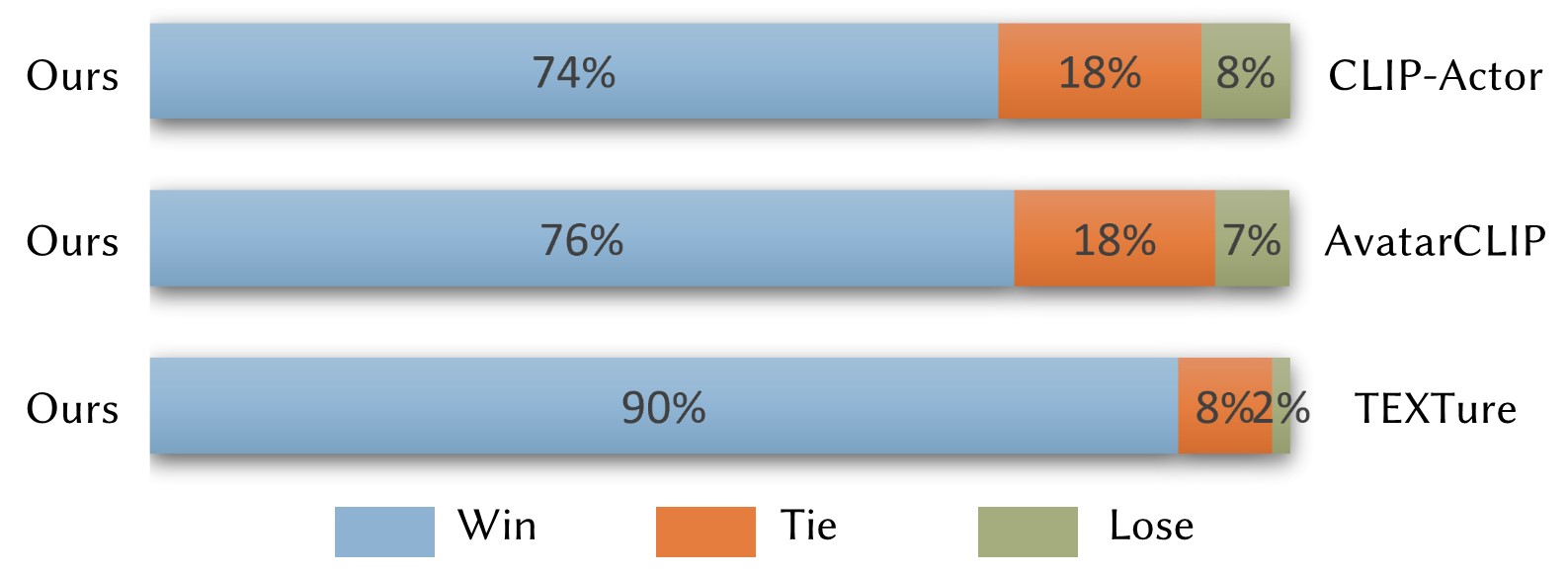}
  \vspace{-0.3in}
  \caption{\textbf{Evaluations by PickScore.} The results of PickScore demonstrate that our method (prompt generative mode) outperforms CLIP-Actor, AvatarCLIP, and TEXTure in visual quality.}  
  \vspace{-0.1in}
\label{fig:pickscore}
\end{figure}

\textbf{Text-to-image metric.} To the best of our knowledge, there are no metrics that can directly and quantitatively evaluate text-to-3D generative models, therefore, we render the generated avatar models to images and then use a text-to-image metric for evaluation.   Specifically, the models generated by our method and previous works are first rendered to $2000$ images from 25 different viewpoints, then the avatar quality is compared by PickScore~\cite{kirstain2023pickapic}, which is a text-to-image metric that gauges the fidelity of generated content based on learned human preferences. As reported in Fig.~\ref{fig:pickscore}, PickScores show that our method outperforms CLIP-Actor, AvatarCLIP, and TEXTure by a large margin, indicating that our results own better subjective quality.

\subsection{Ablation Study}

\begin{figure}[t]
  \centering
  \includegraphics[width=1.0\linewidth]{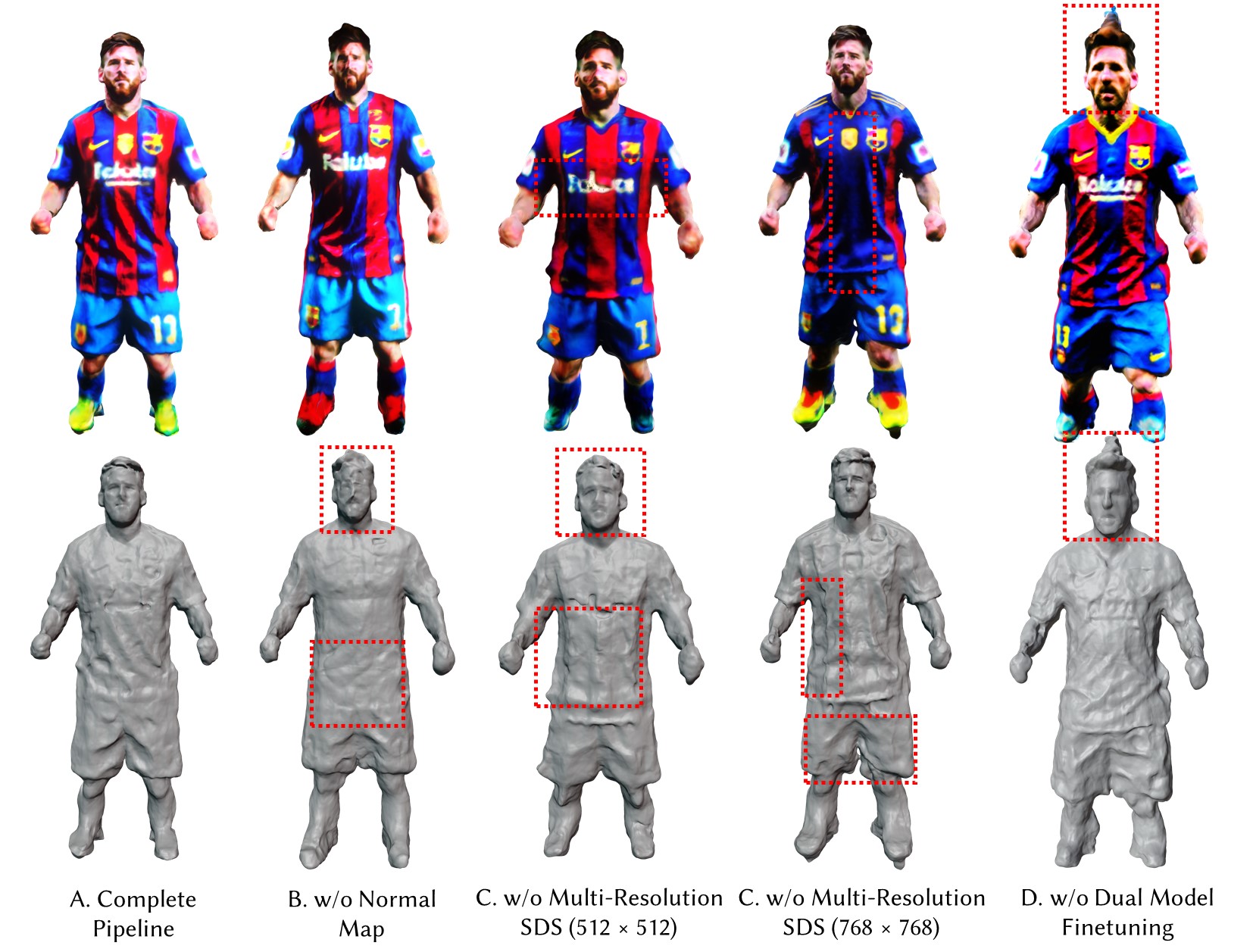}
  \caption{\textbf{Ablation Study-I.} The generated avatars under setting A-D are shown to verify the effectiveness of each module.}
  \label{fig:ablation1}
\end{figure}

To verify the effectiveness of each part of our proposed methods, we conduct ablation experiments to remove certain modules from the complete pipeline, then compare the performance for each setting. The description for each setting is shown below:

\textbf{A. Complete Pipeline.}

\textbf{B. w/o Normal Map (Sec.~\ref{sec:framework})}: The normal maps are not rendered in the training.

\textbf{C. w/o Multi-Resolution SDS (Sec.~\ref{sec:pose_consis})}: The multi-resolution SDS in the complete pipeline is replaced with the strategy of fixed high-resolution SDS.

\textbf{D. w/o Facial Supervision (Sec.~\ref{sec:two_part})}: Only one diffusion model is used in SDS training targeting the full body.

\textbf{E. w/o Pose-Consistent Constraint (Sec.~\ref{sec:pose_consis})}: The pose-consistent constraint is removed from the complete pipeline.

Comparing \textbf{A} and \textbf{B} in Fig.~\ref{fig:ablation1}, we observe that the removal of normal map supervision guidance in setting \textbf{B} leads to a significant loss of geometric details in the generated avatars, while the complete method with normal maps supervision in setting \textbf{A} produces high-quality avatars with improved details, including facial features and clothing wrinkles. These findings support the effectiveness of our proposed normal map guidance strategy in generating realistic geometry from text descriptions.

Comparing \textbf{A} and \textbf{C} in Fig.~\ref{fig:ablation1}, we can see that the avatar generated by a fixed high-resolution upsampling setting contains unclear textures, wrong textures, or unreasonable geometric details, while these issues are eliminated after the multi-resolution SDS is implemented. We think the reason is that the multi-resolution SDS scheme learns multi-scale information in a coarse-to-fine manner, which helps generate a more detailed and clear 3D avatar.

Comparing \textbf{A} and \textbf{D} in Fig.~\ref{fig:ablation1}, it can be seen that the generated face supervised with a single model contains obvious artifacts on the head, and the facial geometry and appearance are less detailed. Besides, the removal of SDS loss for the head leads to an imbalance head-body ratio. By contrast, the avatar face generated under the supervision of dual models contains clear texture and detailed geometry.

\begin{figure}[t]
  \centering
  \includegraphics[width=1.0\linewidth]{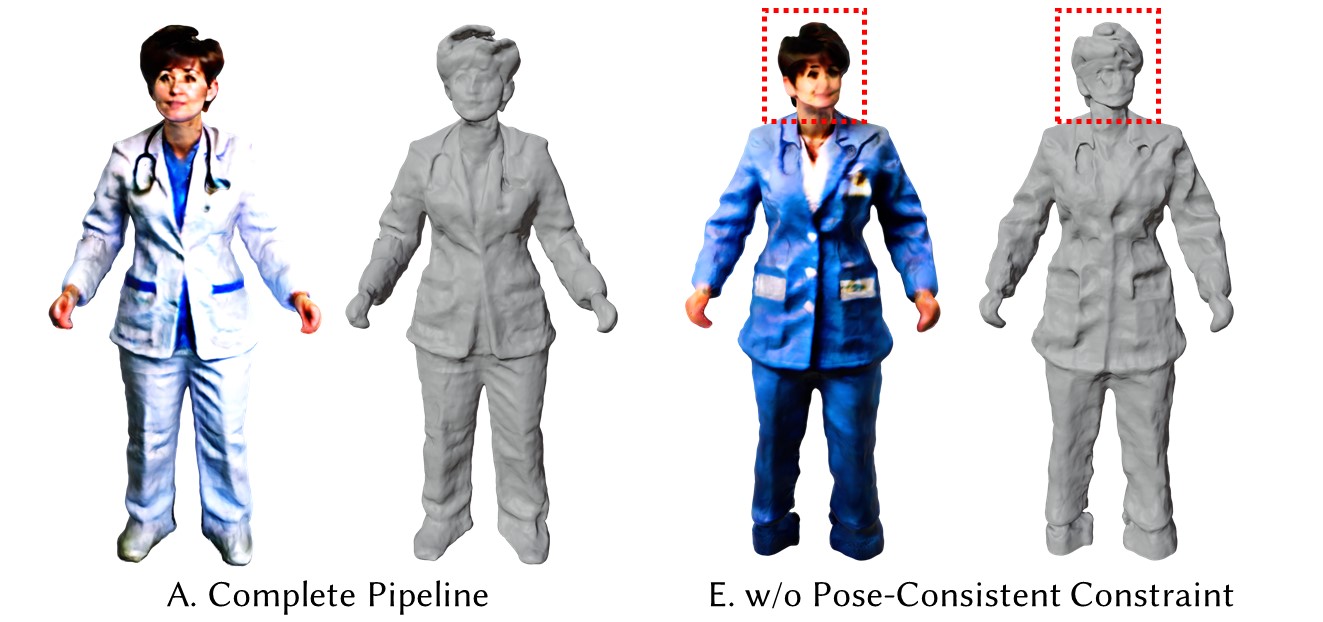}
  \caption{\textbf{Ablation Study-II.} The generated avatars under setting A and E are shown to verify the effectiveness of pose-consistent constraint.}
  \label{fig:ablation2}
\end{figure}

By comparing \textbf{A} and \textbf{E} in Fig.~\ref{fig:ablation2}, we find that the surface of the avatar face tends to converge into a flat or even concave surface if the pose-consistent constraint is removed. We believe that this is because the head poses synthesized by the diffusion model with no pose-consistent constraint are conflicting, which leads to wrong geometry regression under multi-view photometric supervision. In contrast, the introduction of the pose-consistent constraint yields 3D avatars with more plausible geometry.

\section{Conclusion}
\label{sec:con}

In this paper, we propose a method for generating avatar models based on text prompts or free-captured image sets, or both. The human avatar to be synthesized is represented by a neural implicit surface and large pre-trained vision-language models are leveraged for the training of the model via score distillation sampling loss. The pose-consistent constraint is introduced to improve the accuracy of the avatar's geometry and appearance. Dual model fine-tuning and multi-resolution SDS further boost the avatar quality and fidelity in text prompt mode or appearance customized mode. 

There are still some limitations to our approach. The accuracy of the generated models still has the potential to be improved, and the speed of the fine-tuning and training phases still needs to be enhanced. In addition, we do not leverage 3D human datasets for training. Though the data amount of existing 3D human datasets is relatively limited, these priors are still expected to significantly improve the quality of avatar generation.

\noindent\textbf{Acknowledgement.} This work was supported by the NSFC grant 62001213, 62025108, a gift funding from Huawei, and Tencent Rhino-Bird Research Program.

\bibliographystyle{ACM-Reference-Format}
\bibliography{mybib}

\end{document}